\documentclass{article} 
\usepackage{iclr2015,times}
\usepackage{hyperref}
\usepackage{url}

\usepackage{graphicx}
\usepackage{amsmath}
\usepackage{algorithm}
\usepackage{algpseudocode}

\usepackage{mathabx}
\usepackage{textcomp}

\usepackage{amsthm}

\iclrfinalcopy 

\iclrconference 

\newcommand{\bW}{{\mathbf{W}}}

\newcommand{\bu}{{\mathbf{u}}}
\newcommand{\bx}{{\mathbf{x}}}
\newcommand{\by}{{\mathbf{y}}}

\newcommand{\ignore}[1]{}

\title{Embedding Entities and Relations for Learning and Inference in Knowledge Bases}

\author{
Bishan Yang$^{1}$\thanks{Work conducted while interning at Microsoft Research.}, Wen-tau Yih$^{2}$, Xiaodong He$^{2}$, Jianfeng Gao$^{2}$ \& Li Deng$^{2}$\\
$^{1}$Department of Computer Science, Cornell University, Ithaca, NY, 14850, USA\\
\texttt{bishan@cs.cornell.edu} \\
$^{2}$Microsoft Research, Redmond, WA 98052, USA\\
\texttt{\{scottyih,xiaohe,jfgao,deng\}@microsoft.com} \\
}

\begin{document}
\maketitle

\begin{abstract}
We consider learning representations of entities and relations in KBs using the neural-embedding approach. We show that most existing models, including NTN~\citep{SocherChenManningNg2013} and TransE~\citep{BordesUsGaWeYa2013}, can be generalized under a unified learning framework, where entities are low-dimensional vectors learned from a neural network and relations are bilinear and/or linear mapping functions. Under this framework, we compare a variety of embedding models on the link prediction task. We show that a simple bilinear formulation achieves new state-of-the-art results for the task (achieving a top-10 accuracy of 73.2\% vs. 54.7\% by TransE on Freebase). Furthermore, we introduce a novel approach that utilizes the learned relation embeddings to mine logical rules such as $BornInCity(a,b) \land CityInCountry(b,c) \implies Nationality(a,c)$. We find that embeddings learned from the bilinear objective are particularly good at capturing relational semantics, and that the composition of relations is characterized by matrix multiplication. More interestingly, we demonstrate that our embedding-based rule extraction approach successfully outperforms a state-of-the-art confidence-based rule mining approach in mining Horn rules that involve compositional reasoning.
\end{abstract}

\section{Introduction}
Recent years have witnessed a rapid growth of knowledge bases (KBs) such as Freebase\footnote{\url{http://freebase.com}}, DBPedia~\citep{auer2007dbpedia}, and YAGO~\citep{suchanek2007yago}. These KBs store facts about real-world entities (e.g. people, places, and things) in the form of RDF triples\footnote{\url{http://www.w3.org/TR/rdf11-concepts/}} (i.e. (\textit{subject}, \textit{predicate}, \textit{object})). Today's KBs are large in size. For instance, Freebase contains millions of entities and billions of facts (triples) involving a large variety of predicates (relation types). Such large-scale multi-relational data provide an excellent potential for improving a wide range of tasks, from information retrieval, question answering to biological data mining. 

Recently, much effort has been invested in relational learning methods that can scale to large knowledge bases. Tensor factorization (e.g.~\citep{NickelTrKr11,nickel2012factorizing}) and neural-embedding-based models (e.g.~\citep{bordes2013energy,BordesUsGaWeYa2013,SocherChenManningNg2013}) are two popular kinds of approaches that learn to encode relational information using low-dimensional representations of entities and relations. These representation learning methods have shown good scalability and reasoning ability in terms of validating unseen facts given the existing KB.

In this work, we focus on the study of neural-embedding models, where the representations are learned using neural networks with energy-based objectives. Recent embedding models TransE~\citep{BordesUsGaWeYa2013} and NTN~\citep{SocherChenManningNg2013} have shown state-of-the-art prediction performance compared to tensor factorization methods such as RESCAL~\citep{nickel2012factorizing}. They are similar in model forms with slight differences on the choices of entity and relation representations. Without careful comparison, it is not clear how different design choices affect the learning results. In addition, the performance of the embedding models are evaluated on the link prediction task (i.e. predicting the correctness of unseen triples). This only indirectly shows the meaningfulness of low-dimensional embeddings. It is hard to explain what relational properties are being captured and to what extent they are captured during the embedding process.

We make three main contributions in this paper. (1) We present a general framework for multi-relational learning that unifies most multi-relational embedding models developed in the past, including NTN~\citep{SocherChenManningNg2013} and TransE~\citep{BordesUsGaWeYa2013}. (2) We empirically evaluate different choices of entity representations and relation representations under this framework on the canonical link prediction task and show that a simple bilinear formulation achieves new state-of-the-art results for the task (a top-10 accuracy of 73.2\% vs. 54.7\% by TransE when evaluated on Freebase). (3) We propose and evaluate a novel approach that utilizes the learned embeddings to mine logical rules such as $BornInCity(a,b) \land CityOfCountry(b, c) \implies Nationality(a,c)$. We show that such rules can be effectively extracted by modeling the composition of relation embeddings, and that the embeddings learned from the bilinear objective are particularly good at capturing the compositional semantics of relations via matrix multiplication. Furthermore, we demonstrate that our embedding-based approach outperforms a state-of-the-art rule mining system AMIE~\citep{galarraga2013amie} on mining rules that involve compositional reasoning.

The rest of this paper is structured as follows. Section~\ref{sec:related_work} discusses related work. Section~\ref{sec:framework} presents the general framework for learning multi-relational representations. Sections~\ref{sec:link_prediction} and~\ref{sec:rule_extraction} present two inference tasks: a canonical link prediction task and a novel rule extraction task where the learned embeddings are empirically evaluated. Section~\ref{sec:conclusion} concludes the paper.
\section{Related Work}
\label{sec:related_work}
Multi-relational learning has been an active research area for the past couple of years. Traditional statistical learning approaches~\citep{GetoorTa07} such as Markov-logic networks~\citep{richardson2006markov} usually suffer from scalability issues. More recently, various types of representation learning methods have been proposed to embed multi-relational knowledge into low-dimensional representations of entities and relations, including tensor/matrix factorization~\citep{singh2008relational,NickelTrKr11,nickel2012factorizing}, Bayesian clustering framework~\citep{kemp2006learning, sutskever2009modelling}, and neural networks~\citep{paccanaro2001learning,bordes2013energy,BordesUsGaWeYa2013,SocherChenManningNg2013}. Our work focuses on the study of neural-embedding models as they have shown good scalability and strong generalizability on large-scale KBs.

Existing neural embedding models~\citep{bordes2013energy,BordesUsGaWeYa2013,SocherChenManningNg2013} all represent entities as low-dimensional vectors and represent relations as operators that combine the representations of two entities. They differ in different parametrization of relation operators. For instance, given two entity vectors, the model of Neural Tensor Network (NTN)~\citep{SocherChenManningNg2013} represents each relation as a bilinear tensor operator followed by a linear matrix operator. The model of TransE~\citep{BordesUsGaWeYa2013}, on the other hand, represents each relation as a single vector that linearly interacts with the entity vectors. Likewise, variations on entity representations also exist. Most methods represent each entity as a unit vector while NTN~\citep{SocherChenManningNg2013} represent entities as an average of word vectors and initializing word vectors with pre-trained vectors from external text corpora. There has not been work that closely examines the effectiveness of these different design choices. 

Our work on embedding-based rule extraction presented in part of this paper is related to the earlier study on logical inference with learned continuous-space representations. Much existing work along this line focuses on learning logic-based representations for natural language sentences. For example,~\citet{socher:semantic} builds a neural network that recursively combines word representations based on parse tree structures and shows that such neural network can simulate the behavior of conjunction and negation. \citet{Bowman2014} further demonstrates that recursive neural network can capture certain aspects of natural logical reasoning on examples involving quantifiers like \textit{some} and \textit{all}. Recently, \citet{Grefenstette2013} shows that in theory most aspects of predicate logic can be simulated using tensor calculus. \citet{Tim2014} further implements the idea by introducing a supervised objective that trains embeddings to be consistent with given logical rules. The evaluation was conducted on toy data and uses limited logical forms. Different from these earlier studies, we propose a novel approach to utilizing embeddings learned without explicit logical constraints to directly mine logical rules from KBs. We demonstrate that the learned embeddings of relations can capture the compositional semantics of relations. Moreover, we systematically evaluate our approach and compare it favorably with a state-of-the-art rule mining approach on the rule extraction task on Freebase.

%
\section{Multi-Relational Representation Learning}
\label{sec:framework}
In this section, we present a general neural network framework for multi-relational representation learning. We discuss different design choices for the representations of entities and relations which will be empirically compared in Section~\ref{sec:link_prediction}.

Given a KB that is represented as a list of relation triplets $(e_1,r,e_2)$ (denoting $e_1$ (the~\textit{subject}) and $e_2$ (the~\textit{object}) that are in a certain relationship $r$), we want to learn representations for entities and relations such that valid triplets receive high scores (or low energies). The embeddings can be learned via a neural network. The first layer projects a pair of input entities to low dimensional vectors, and the second layer combines these two vectors to a scalar for comparison via a scoring function with relation-specific parameters. 

\subsection{Entity Representations}
Each input entity corresponds to a high-dimensional vector, either a ``one-hot" index vector or a ``n-hot" feature vector. Denote by $\bx_{e_1}$ and $\bx_{e_2}$ the input vectors for entity $e_1$ and $e_2$, respectively. Denote by $W$ the first layer projection matrix. The learned entity representations, $\by_{e_1}$ and $\by_{e_2}$ can be written as
$$\by_{e_1}=f\big(\bW \bx_{e_1}\big),~~ \by_{e_2}=f\big(\bW \bx_{e_2}\big)$$
where $f$ can be a linear or non-linear function, and $\bW$ is a parameter matrix, which can be randomly initialized or initialized using pre-trained vectors. 

Most existing embedding models adopt the ``one-hot" input vectors except for NTN~\citep{SocherChenManningNg2013} which represents each entity as an average of its word vectors. This can be viewed as adopting ``bag-of-words" vectors as input and learning a projection matrix consisting of word vectors. 
\subsection{Relation Representations}
The choice of relation representations reflects in the form of the scoring function. Most of the existing scoring functions in the literature can be unified based on a basic linear transformation $g_r^a$, a bilinear transformation $g_r^b$ or their combination, where $g_r^a$ and $g_r^b$ are defined as
\begin{equation}
g_r^a(\by_{e_1},\by_{e_2})=\mathbf{A}_r^T\left(\begin{array}{c}\by_{e_1}\\ \by_{e_2}\end{array}\right)~~\textrm{and} ~~~g_r^b(\by_{e_1},\by_{e_2})=\by_{e_1}^T \mathbf{B}_r \by_{e_2},
\label{g}
\end{equation}
which $\mathbf{A}_r$ and $\mathbf{B}_r$ are relation-specific parameters.

\begin{table*}[bth]
\begin{center}
\begin{footnotesize}
\scalebox{0.9}{
\begin{tabular}{|c|c|c|c|}
\hline
Models & $\mathbf{B}_r$ & $\mathbf{A}_r^T$ & Scoring Function\\
\hline
Distance~\citep{bordes2011learning} & - & $\big(\mathbf{Q}_{r_1}^T\;~\;{-\mathbf{Q}_{r_2}^T}\big)$ & $-||g_r^a(\by_{e_1}, \by_{e_2})||_1$\\
\hline
Single Layer~\citep{SocherChenManningNg2013} & - & $\big(\mathbf{Q}_{r1}^T\;~~~~\;\mathbf{Q}_{r2}^T\big)$ & $\bu_r^T \tanh(g_r^a(\by_{e_1},\by_{e_2}))$\\
\hline
TransE~\citep{BordesUsGaWeYa2013} & $\mathbf{I}$ & $\big(\mathbf{V}_r^T\;~\;{-\mathbf{V}_r^T}\big)$ & $-(2g_r^a(\by_{e_1}, \by_{e_2})-2g_r^b(\by_{e_1}, \by_{e_2})+||\mathbf{V}_r||_2^2)$\\
\hline
NTN~\citep{SocherChenManningNg2013} & $\mathbf{T}_r$ & $\big(\mathbf{Q}_{r1}^T\;\;~~~~\mathbf{Q}_{r2}^T\big)$ & $\bu_r^T \tanh\big(g_r^a(\by_{e_1},\by_{e_2})+g_r^b(\by_{e_1},\by_{e_2})\big)$\\
\hline
\end{tabular}
}
\end{footnotesize}
\caption{\label{summary} Comparisons among several multi-relational models in their scoring functions. }
\end{center}
\end{table*}
%

In Table~\ref{summary}, we summarize several popular scoring functions in the literature for a relation triplet~$(e_1,r,e_2)$, reformulated in terms of the above two functions. Denote by $\by_{e_1},\by_{e_2}\in \it{R}^n$ two entity vectors. 
Denote by $\mathbf{Q}_{r_1}, \mathbf{Q}_{r_2}\in \it{R}^{n\times m}$ and $\mathbf{V}_r\in \it{R}^n$ matrix or vector parameters for linear transformation $g_r^a$ . 
Denote by $\mathbf{T}_r\in \it{R}^{n\times n\times m}$ tensor parameters for bilinear transformation $g_r^b$. $\mathbf{I}\in \it{R}^n$ is an identity matrix. $\bu_r \in \it{R}^{m}$ is an additional parameter for relation $r$. The scoring function for TransE (L2 formulation) is derived from $||\by_{e_1}-\by_{e_2}+V_r||_2^2=2V_r^T(\by_{e_1}-\by_{e_2})-2\by_{e_1}^T\by_{e_2}+||V_r||_2^2+||\by_{e_1}||_2^2+||\by_{e_2}||_2^2$, where $\by_{e_1}$ and $\by_{e_2}$ are unit vectors. 

Note that NTN is the most expressive model as it contains both linear and bilinear relation operators as special cases. In terms of the number of parameters, TransE is the simplest model which only parametrizes the linear relation operators with one-dimensional vectors. 

In this paper, we also consider the basic bilinear scoring function:
\begin{equation}
\label{bilinear}
g_r^b(\by_{e_1},\by_{e_2})=\by_{e_1}^T\mathbf{M}_r\by_{e_2}
\end{equation} 
which is a special case of NTN without the non-linear layer and the linear operator, and uses a 2-d matrix operator $\mathbf{M}_r\in \it{R}^{n \times n}$ instead of a tensor operator. Such bilinear formulation has been used in other matrix factorization models such as in ~\citep{NickelTrKr11,jenatton2012latent,garcia2014effective} with different forms of regularization. Here, we consider a simple way to reduce the number of relation parameters by restricting $\mathbf{M}_r$ to be a diagonal matrix. This results in the same number of relation parameters as TransE. Our experiments in Section~\ref{sec:link_prediction} demonstrate that this simple formulation enjoys the same scalable property as TransE and it achieves superior performance over TransE and other more expressive models on the task of link prediction.

This general framework for relationship modeling also applies to the recent deep-structured semantic model~\citep{Huang-2013,Shen-2014,shen2014learning,Gao-2014,yihsemantic}, which learns the relevance or a single relation between a pair of word sequences. The framework above applies when using multiple neural network layers to project entities and using a relation-independent scoring function $G_r\big(\by_{e_1}, \by_{e_2}\big)=\cos[\by_{e_1}(\mathbf{W}_r),\by_{e_2}(\mathbf{W}_r)]$. The cosine scoring function is a special case of $g_r^b$ with normalized $\by_{e_1},\by_{e_2}$ and with $\mathbf{B}_r=\mathbf{I}$.

\subsection{Parameter Learning}
The neural network parameters of all the models discussed above can be learned by minimizing a margin-based ranking objective
, which encourages the scores of positive relationships (triplets) to be higher than the scores of any negative relationships (triplets). Usually only positive triplets are observed in the data. Given a set of positive triplets $T$, we can construct a set of ``negative" triplets $T'$ by corrupting either one of the relation arguments, $T'=\{(e_1',r,e_2)|e_1'\in E,(e_1',r,e_2)\notin T\}\cup \{(e_1,r,e_2')|e_2'\in E, (e_1,r,e_2')\notin T\}$. Denote the scoring function for triplet $(e_1,r,e_2)$ as $S_{(e_1, r, e_2)}$. The training objective is to minimize the margin-based ranking loss
\begin{equation}
\label{obj}
L(\Omega)=\sum_{(e_1,r,e_2)\in T}\sum_{(e_1',r,e_2')\in T'}\max\{S_{(e_1',r,e_2')}-S_{(e_1,r,e_2)}+1, 0\}
\end{equation}
\section{Inference Task I: Link Prediction}
\label{sec:link_prediction}
We first conduct a comparison study of different embedding models on the canonical link prediction task, which is to predict the correctness of unseen triplets. As in~\citep{BordesUsGaWeYa2013}, we formulate link prediction as an entity ranking task. For each triplet in the test data, we treat each entity as the target entity to be predicted in turn. Scores are computed for the correct entity and all the corrupted entities in the dictionary and are ranked in descending order. We consider \textit{Mean Reciprocal Rank~(MRR)} (an average of the reciprocal rank of an answered entity over all test triplets), \textit{HITS@10} (top-10 accuracy), and \textit{Mean Average Precision}~(MAP) (as used in~\citep{CYYM14}) as the evaluation metrics.

We examine five embedding models in decreasing order of complexity: (1) NTN with $4$ tensor slices as in~\citep{SocherChenManningNg2013}; (2) Bilinear+Linear, NTN with $1$ tensor slice and without the non-linear layer; (3) TransE, a special case of Bilinear+Linear (see Table~\ref{summary}); (4) Bilinear: using scoring function in Eq.~\eqref{bilinear}; (5) Bilinear-diag: a special case of Bilinear where the relation matrix is a diagonal matrix.

\paragraph{Datasets} We used the WordNet (WN) and Freebase (FB15k) datasets introduced in~\citep{BordesUsGaWeYa2013}. WN contains $151,442$ triplets with $40,943$ entities and $18$ relations, and FB15k consists of $592,213$ triplets with $14,951$ entities and $1345$ relations. We use the same training/validation/test split as in~\citep{BordesUsGaWeYa2013}. We also consider a subset of FB15k (FB15k-401) containing only frequent relations (relations with at least $100$ training examples). This results in $560,209$ triplets with $14,541$ entities and $401$ relations.  

\paragraph{Implementation details} All the models were implemented in C\# and using GPU. Training was implemented using mini-batch stochastic gradient descent with AdaGrad~\citep{duchi2011adaptive}. At each gradient step, we sampled for each positive triplet two negative triplets, one with a corrupted subject entity and one with a corrupted object entity. The entity vectors are renormalized to have unit length after each gradient step (it is an effective technique that empirically improved all the models). For the relation parameters, we used standard L2 regularization. For all models, we set the number of mini-batches to $10$, the dimensionality of the entity vector $d=100$, the regularization parameter $0.0001$, and the number of training epochs $T=100$ on FB15k and FB15k-401 and $T=300$ on~WN ($T$ was determined based on the learning curves where the performance of all models plateaued.) The learning rate was initially set to $0.1$ and then adapted during training by AdaGrad. 

\subsection{Results}
\begin{table*}[bth]
\begin{center}
\begin{footnotesize}
\scalebox{0.9}{
\begin{tabular}{|c|c|c|c|c|c|c|}
\hline
& \multicolumn{2}{|c|}{FB15k}& \multicolumn{2}{|c|}{FB15k-401}& \multicolumn{2}{|c|}{WN}\\
& MRR & HITS$@$10 & MRR & HITS$@$10 & MRR & HITS$@$10\\
\hline
NTN & 0.25 & 41.4 & 0.24 & 40.5 & 0.53 & 66.1 \\
\hline
Blinear+Linear & 0.30 & 49.0 & 0.30 & 49.4 & 0.87 & 91.6\\
\hline
TransE ({\sc DistADD}) & 0.32 & 53.9 & 0.32 & 54.7 & 0.38 & 90.9\\
\hline
Bilinear & 0.31 & 51.9 & 0.32 & 52.2 & \textbf{0.89} & 92.8 \\
\hline
Bilinear-diag ({\sc DistMult}) & \textbf{0.35} & \textbf{57.7} & \textbf{0.36} & \textbf{58.5} & 0.83 & \textbf{94.2} \\
\hline
\end{tabular}
}
\end{footnotesize}
\caption{\label{comparison}Performance comparisons among different embedding models}
\end{center}
\end{table*}

Table~\ref{comparison} shows the results of all compared methods on all the datasets. In general, we observe that the performance increases as the complexity of the model decreases on FB. NTN, the most complex model, provides the worst performance on both FB and WN, which suggests overfitting. Compared to the previously published results of TransE~\citep{BordesUsGaWeYa2013}, our implementation achieves much better results~(53.9\% vs. 47.1\% on FB15k and 90.9\% vs. 89.2\% on WN) using the same evaluation metric~(HITS@10). We attribute such discrepancy mainly to the different choice of SGD optimization: AdaGrad vs. constant learning rate. We also found that Bilinear consistently provides comparable or better performance than TransE, especially on WN. Note that WN contains much more entities than FB, it may require the parametrization of relations to be more expressive to better handle the richness of entities. Interestingly, we found that a simple variant of Bilinear -- {\sc Bilinear-diag}, clearly outperforms all baselines on FB and achieves comparable performance to Bilinear on WN. Note that {\sc Bilinear-diag} has the limitation of encoding the difference between a relation and its inverse. Still, as there is a large variety of relations in FB and the average number of training examples seen by each relation is relatively small (compared to WN), the simple form of {\sc Bilinear-diag} is able to provide good prediction performance. 

\textbf{Multiplicative vs. Additive Interactions} Note that {\sc Bilinear-diag} and {\sc TransE} have the same number of model parameters and their difference can be viewed as the operational choices of the composition of two entity vectors -- {\sc Bilinear-diag} uses weighted element-wise dot product (multiplicative operation) and {\sc TransE} uses element-wise subtraction with a bias (additive operation). To highlight the difference, here we use {\sc DistMult} and {\sc DistAdd} to refer to {\sc Bilinear-diag} and {\sc TransE}, respectively.
Comparisons between these two models can provide us more insights on the effect of two common choices of compositional operations -- multiplication and addition for modeling entity relations. Overall, we observed superior performance of {\sc DistMult} on all the datasets in Table~\ref{comparison}. 
Table~\ref{details} shows the \textit{HITS@10} score on four types of relation categories (as defined in~\citep{BordesUsGaWeYa2013}) on FB15k-401 when predicting the subject entity and the object entity respectively. We can see that {\sc DistMult} significantly outperforms {\sc DistAdd} in almost all the categories. 
\begin{table*}[bth]
\begin{center}
\begin{footnotesize}
\scalebox{0.9}{
\begin{tabular}{|c|c|c|c|c|c|c|c|c|}
\hline
& \multicolumn{4}{|c|}{Predicting subject entities}& \multicolumn{4}{|c|}{Predicting object entities}\\
\hline
& 1-to-1 & 1-to-n & n-to-1 & n-to-n & 1-to-1 & 1-to-n & n-to-1 & n-to-n\\
\hline
{\sc DistADD} & 70.0 & 76.7 & 21.1 & 53.9 & 68.7 & 17.4 &~\textbf{83.2} & 57.5\\
\hline
{\sc DistMult} & \textbf{75.5} & \textbf{85.1} & \textbf{42.9} & \textbf{55.2} & \textbf{73.7} & \textbf{46.7} & 81.0 & \textbf{58.8}\\
\hline
\end{tabular}
}
\end{footnotesize}
\caption{\label{details}Results by relation categories: one-to-one, one-to-many, many-to-one and many-to-many}
\end{center}
\end{table*}

\textbf{Initialization of Entity Vectors} In the following, we examine the learning of entity representations and introduce two further improvements: using non-linear projection and initializing entity vectors with pre-trained vectors. We focus on {\sc DistMult} as our baseline and compare it with the two modifications {\sc DistMult}-tanh (using $f=\tanh$ for entity projection
) and {\sc DistMult}-tanh-EV-init (initializing the entity parameters with the $1000$-dimensional pre-trained entity vectors released by \textit{word2vec}~\citep{mikolov2013distributed}) on FB15k-401. We also reimplemented the initialization technique introduced in~\citep{SocherChenManningNg2013} -- each entity is represented as an average of its word vectors and the word vectors are initialized using the $300$-dimensional pre-trained word vectors released by \textit{word2vec}. We denote this method as {\sc DistMult}-tanh-WV-init. Inspired by~\citep{CYYM14}, we design a new evaluation setting where the predicted entities are automatically filtered according to ``entity types" (entities that appear as the subjects/objects of a relation have the same type defined by that relation). This provides us with better understanding of the model performance when some entity type information is provided. 
\begin{table*}[bth]
\begin{center}
\scalebox{0.8}{
\begin{tabular}{|c|c|c|c|}
\hline
& MRR & HITS$@$10 & MAP (w/ type checking)\\
\hline
{\sc DistMult} & 0.36 & 58.5 & 64.5\\
\hline
{\sc DistMult}-tanh & 0.39 & 63.3 & 76.0\\
\hline
{\sc DistMult}-tanh-WV-init & 0.28 & 52.5 & 65.5\\
\hline
{\sc DistMult}-tanh-EV-init & \textbf{0.42} & \textbf{73.2} & \textbf{88.2}\\
\hline
\end{tabular}
}
\caption{\label{further_exp}Evaluation with pre-trained vectors}
\end{center}
\end{table*}

In Table~\ref{further_exp}, we can see that {\sc DistMult}-tanh-EV-init provides the best performance on all the metrics. Surprisingly, we observed performance drops by {\sc DistMult}-tanh-WV-init. We suspect that this is because word vectors are not appropriate for modeling entities described by non-compositional phrases (more than 73\% of the entities in FB15k-401 are person names, locations, organizations and films). The promising performance of {\sc DistMult}-tanh-EV-init suggests that the embedding model can greatly benefit from pre-trained entity-level vectors using external textual resources. 

\section{Inference Task II: Rule Extraction}
\label{sec:rule_extraction}
In this section, we focus on a complementary inference task, where we utilize the learned embeddings to extract logical rules from the KB. For example, given the fact that a person was born in New York and New York is a city of the United States, then the person's nationality is the United States:
\[BornInCity(a,b) \land CityOfCountry(b, c) \implies Nationality(a,c)\]
Such logical rules can serve four important purposes. First, they can help deduce new facts and complete the existing KBs. Second, they can help optimize data storage by storing only rules instead of large amounts of extensional data, and generate facts only at inference time. Third, they can support complex reasoning. Lastly, they can provide explanations for inference results, e.g. we may infer that people's professions usually involve the specialization of the field they study, etc.

The key problem of extracting Horn rules like the aforementioned example is how to effectively explore the search space. Traditional rule mining approaches directly operate on the KB graph -- they search for possible rules (i.e. closed-paths in the graph) by pruning rules with low statistical significance and relevance~\citep{schoenmackers2010learning}. These approaches often fail on large KB graphs due to scalability issues. In the following, we introduce a novel embedding-based rule mining approach whose efficiency is not affected by the size of the KB graph but rather by the number of distinct types of relations in the KB (which is usually relatively small). It can also mine better rules due to its strong generalizability.

\subsection{Background and Notations} 
For a better illustration, we adopt the graph view of KB. Each binary relation $r(a,b)$ is a directed edge from node $a$ to node $b$ and with link type $r$. We are interested in extracting Horn rules that consist of a \textbf{head} relation $H$ and a sequence of \textbf{body} relations $B_1,...,B_n$:
\begin{equation}
\label{horn_rule}
B_1(a_1,a_2)\land B_2(a_2,a_3) \land ... \land B_n(a_n, a_{n+1}) \implies H(a_1,a_{n+1})
\end{equation}
where $a_i$ are variables that can be substituted by entities. We constrain the body relations $B_1,...,B_n$ to form a directed \textit{path} in the graph and the head relation $H$ to from a directed edge that \textit{close} the path (from the start of the path to the end of the path). We denote such property as the \textbf{closed-path} property. For consecutive relations that share one variable but do not form a path, e,g, $B_{i-1}(a, b) \land B_i(a, c)$, we can replace one of the relations with its inverse relation, so that the relations are connected by an object and an subject, e.g. $B^{-1}_{i-1}(b, a) \land B_i(a, c)$. We are interested in mining rules that reflect relationships among different relation types, therefore we also constrain $B_1,...,B_n,H$ to have distinct relation types. A rule is \textbf{instantiated} when all variables are substituted by entities. We denote the \textbf{length} of the rule as the number of body relations. In general longer rules are harder to extract due to the exponential search space. In our experiments, we focus on extracting rules of length 2 and 3.

In KBs, entities usually have types and relations often can only take arguments of certain types. For example, \textit{BornInCity} relation can only take a \textit{person} as the subject and a \textit{location} as the object. For each relation $r$, we denote the domain of its subject argument (the set of entities that can appear in the subject position) as $\mathcal{X}_r$ and similarly the domain of its object argument as $\mathcal{Y}_r$. Such domain information can be extremely useful in restricting the search space of logical rules.

\subsection{Embedding-based Rule Extraction}
\label{sec:embedrule}
For simplicity, we consider Horn rules of length 2 (longer rules can be easily derived from this case):
\begin{equation}
\label{3-atom-rule}
B_1(a,b) \land B_2(b,c) \implies H(a,c)
\end{equation}
Note that the body of the rule can be viewed as the composition of relations $B_1$ and $B_2$, which is a new relation that has the property that entities $a$ and $c$ are in a relation if and only if there is an entity $b$ which simultaneously satisfies two relations $B_1(a,b)$ and $B_2(b,c)$. 

We model relation composition as multiplication or addition of two relation embeddings. Here we focus on relation embeddings that are in the form of vectors (as in {\sc TransE}) and matrices (as in {\sc Bilinear} and its variants). The composition results in a new embedding that lies in the same relation space. Specifically, we use addition for relation vector embeddings and multiplication for relation matrix embeddings. This is inspired by two different properties: (1) if a relation corresponds to a translation vector $V$ and assume $\by_a + \mathbf{V}-\by_b\approx0$ when $B(a,b)$ holds, then we have the property that $\by_a + \mathbf{V}_1\approx\by_b$ and $\by_b + \mathbf{V}_2\approx\by_c$ implies $\by_a + (\mathbf{V}_1\ + \mathbf{V}_2)\approx\by_c$; (2) if a relation corresponds to a matrix $M$ in the bilinear transformation and assume $\by_a^T\mathbf{M}\by_b\approx1$ when $B(a,b)$ holds, also $\by_a$ and $\by_b$ are unit vectors and $\by_a^T\mathbf{M}$ is still a unit vector~\footnote{These assumptions may not hold in our implementations. The intuition still leads to surprisingly good empirical performance on Horn rule extraction. How to effectively enforce these constraints is worth investigating in future work.}, then we have the property that $\by_a^T\mathbf{M}_1\approx\by_b^T$ and $\by_b^T\mathbf{M}_2\approx\by_c^T$ implies $\by_a^T(\mathbf{M}_1\mathbf{M}_2)\approx\by_c^T$.

To simulate the implication in~\ref{3-atom-rule}, we want the composition result of relation $B_1$ and $B_2$ to demonstrate similar behavior to the embedding of relation $H$. We assume the similarity between relation embeddings is related to the Euclidean distance if the embeddings are vectors and to the Frobenius norm if the embeddings are matrices. This distance metric allows us to rank possible pairs of relations with respect to the relevance of their composition to the target relation. 

Note that we do not need to enumerate all possible pairs of relations in the KB. For example, if the relation in the head is $r$, then we are only interested in relation pairs $(p,q)$ that satisfy the type constraints, namely: (1) $\mathcal{Y}_{p}\cap \mathcal{X}_{q}\neq \emptyset$; (2) $\mathcal{X}_{p}\cap \mathcal{X}_{r}\neq \emptyset$; (3) $\mathcal{Y}_{q}\cap \mathcal{Y}_{r}\neq \emptyset$. As mentioned before, the arguments (entities) of relations are usually strongly typed in KBs. Applying such domain constraints can effectively reduce the search space. 

\begin{algorithm}
\caption{{\sc EmbedRule}}
\label{rule_extraction_alg}
\begin{algorithmic}[1]
\State \textbf{Input:} $KB=\{(e_1,r,e_2)\}$, relation set $R$
\State \textbf{Output:} Candidate rules $Q$
\For{\textbf{each} $r$ in $R$}
    \State Select the set of start relations $S=\{s:\mathcal{X}_{s}\cap \mathcal{X}_{r}\neq \emptyset\}$ 
    \State Select the set of end relations $T=\{t:\mathcal{Y}_{t}\cap \mathcal{Y}_{r}\neq \emptyset\}$
    \State Find all possible relation sequences 
    \label{algm:EmbedRule:seq}
    \State Select the $K$-NN sequences $P' \subseteq P$ for $r$ based on $dist(\mathbf{M}_r,\mathbf{M}_{p_1}\circ\cdots\circ\mathbf{M}_{p_n})$ \label{algm:EmbedRule:k-NN}    
    \State Form candidate rules using $P'$ where $r$ is the head relation and $p \in P'$ is the body in a rule
    \State Add the candidate rules into $Q$
\EndFor
\end{algorithmic}
\label{algm:EmbedRule}
\end{algorithm}

In Algorithm~\ref{algm:EmbedRule}, we describe our rule extraction algorithm for general closed-path Horn rules as in Eq.~\eqref{horn_rule}. In Step~\ref{algm:EmbedRule:k-NN}, $\circ$ denotes vector addition or matrix multiplication. We apply a global threshold value $\delta$ in our experiments to filter candidate sequences for each relation $r$, and then automatically select the top remaining sequences by applying a heuristic thresholding strategy based on the difference of the distance scores: sort the sequences by increasing distance $d_1,...,d_{T}$ and set the cut-off point to be the $j$-th sequence where $j=\arg\max_{i}(d_{i+1}-d_{i})$.  

\subsection{Experiments} 
We evaluate our rule extraction method (denoted as {\sc EmbedRule}) on the FB15k-401 dataset. In our experiments, we remove the equivalence relations and relations whose domains have cardinality $1$ since rules involving these relations are not interesting. This results in training data that contains 485,741 facts, 14,417 entities, and 373 relations. Our {\sc EmbedRule} algorithm identifies 60,020 possible length-2 relation sequences and 2,156,391 possible length-3 relation sequences. We then apply the thresholding method described in Section~\ref{sec:embedrule} to further select top $\sim$3.9K length-2 rules and $\sim$2K length-3 rules~\footnote{We consider $K$=100 nearest-neighbor sequences for each method, and set $\delta$ to 9.2, 36.3, 1.9 and 3.4 for {\sc DistMult-tanh-EV-init}, {\sc DistMult}, {\sc Bilinear} and {\sc DistAdd} respectively for length-2 rules, and set it to 9.1, 48.8, 2.9, and 1.1 for lengh-3 rules.}. By default all the extracted rules are ranked by decreasing confidence, which is computed as the ratio of the correct predictions to the total number of predictions, where predictions are triplets that are derived from the instantiated rules where the body relations are observed. 

We implemented four versions of {\sc EmbedRule} using embeddings trained from {\sc TransE (DistAdd)}, {\sc Bilinear}, {\sc Bilinear-diag (DistMult)} and {\sc DistMult}-tanh-EV-init with corresponding composition functions. We also compare our approaches to {\sc AMIE}~\citep{galarraga2013amie}, a state-of-the-art rule mining system that can efficiently search for Horn rules in large-scale KBs by using novel measurements of support and confidence. The system extracts \textit{close} rules -- a superset of the rules we consider in this paper: every relation in the body is connected to the following relation by sharing an entity variable, and every entity variable in the rule appears \textit{at least} twice. We run AMIE with the default setting on the same training set. It extracts 2,201 possible length-1 rules and 46,975 possible length-2 rules, among which 3,952 rules have the \textit{close-path} property. We compare these length-2 rules with the similar number of length-2 rules extracted by {\sc EmbedRule}. By default AMIE ranks rules by PCA confidence (a normalized confidence that takes into account the incompleteness of KBs). However we found that ranking by the standard confidence gives better performance than the PCA confidence on the Freebase dataset we use. 

For computational cost, ${\sc EmbedRule}$ mines length-2 rules in $2$ minutes and mines length-3 rules in $20$ minutes (the computational time is similar when using different types of embeddings). {\sc AMIE} mines rules of length $\leq2$ in $9$ minutes. All methods are evaluated on a machine with a 64-bit processor, 2 CPUs and 8GB memory.

We consider precision as the evaluation metric, which is the ratio of predictions that are in the test (unseen) data to all the generated unseen predictions. Note that this is an estimation, since a prediction is not necessarily ``incorrect" if it is not seen. \citet{galarraga2013amie} suggested to identify incorrect predictions based on the functional property of relations. However, we find that most relations in our datasets are not functional. For a better estimation, we manually labeled the top 30 unseen facts predicted by each method by checking Wikipedia. We also remove rules where the head relations are hard to justified due to dynamic factors (i.e. involving the word ``current").
%
%
%
\subsection{Results}
\begin{figure}[ht]
\centering
\includegraphics[width=.5\linewidth]{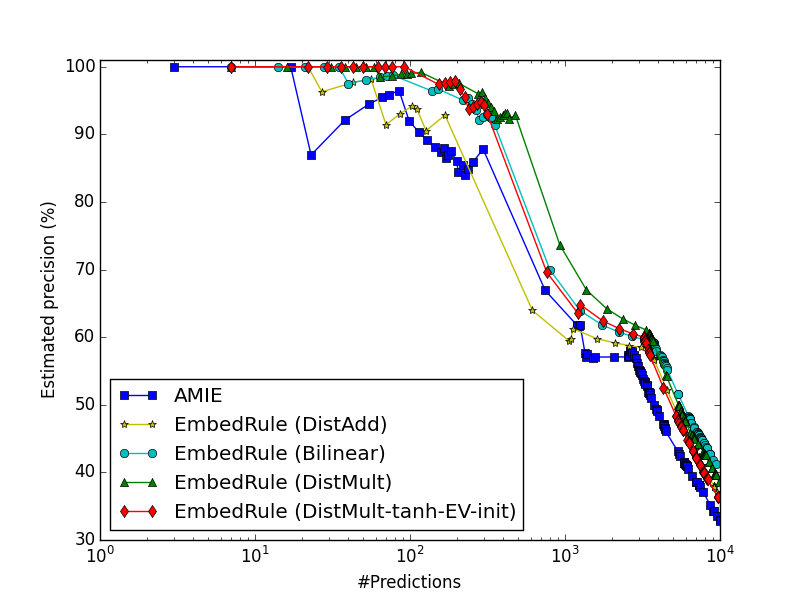}
\caption{Aggregated precision of top length-2 rules extracted by different methods}
\label{fig:aggregate-precision}
\end{figure}

Figure~\ref{fig:aggregate-precision} compares the predictions generated by the length-2 rules extracted by different methods. We plot the aggregated precision of the top rules that produce up to $10K$ predictions in total. From left to right, the $n$-th data point represents the total number of predictions of the top $n$ rules and the estimated precision of these predictions. We can see that {\sc EmbedRule} that uses embeddings trained from the bilinear objective ({\sc Bilinear}, {\sc DistMult} and {\sc DistMult-tanh-EV-init}) consistently outperforms {\sc AMIE}. This suggests that the bilinear embeddings contain good amount of information about relations which allows for effective rule selection without looking at the entities. For example, {\sc AMIE} fails to extract $TVProgramCountryofOrigin(a, b) \land CountryOfficialLanguage(b, c) \implies TVProgramLanguage(a, c)$ by relying on the instantiations of the rule occurred in the observed KB while all the bilinear variants of {\sc EmbedRule} successfully extract the rule purely based on the embeddings of the three involved relations. 

We can also see that in general, using multiplicative composition of matrix embeddings (from {\sc DistMult} and {\sc Bilinear}) results in better performance compared to using additive composition of vector embeddings (from {\sc DistAdd}). We found many examples where {\sc DistAdd} fails to retrieve rules because it assigns large distance between the composition of the body relations and the head relation in the embedding space while its multiplicative counterpart {\sc DistMult} ranks the composition result much closer to the head relation. For example, {\sc DistAdd} prunes the possible composition $FilmDistributorInRegion \land RegionGDPCurrency$ for relation $FilmBudgetCurrency$ while {\sc DistMult} ranks the composition as the nearest neighbor of $FilmBudgetCurrency$. 

\begin{figure}[ht]
\centering
\includegraphics[width=.5\linewidth]{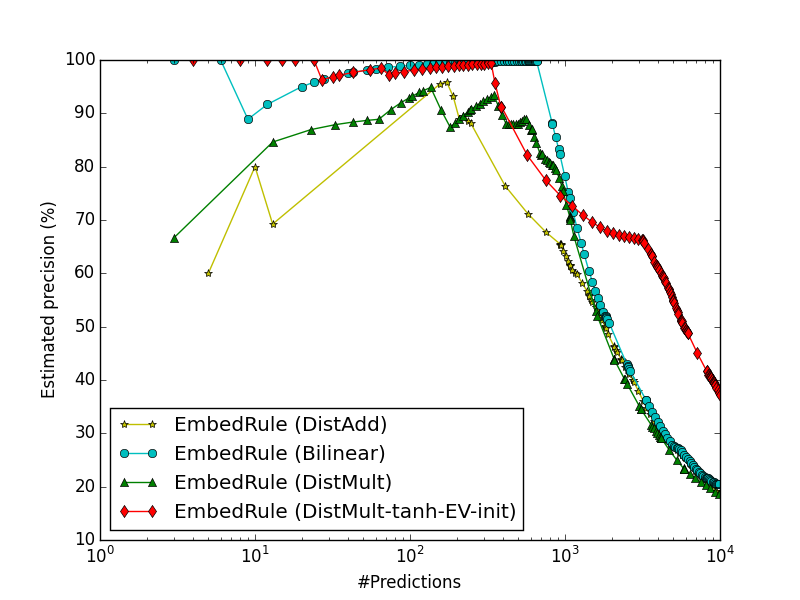}
\caption{Aggregated precision of top length-3 rules extracted by different methods}
\label{fig:precision-len-3}
\end{figure}
We also look at the results for length-3 rules generated by different implementations of {\sc EmbedRule} in Figure~\ref{fig:precision-len-3}. We can see that the initial length-3 rules extracted by {\sc EmbedRule} can provide very good precision in general. We can also see that {\sc Bilinear} consistently outperforms {\sc DistMult} and {\sc DistAdd} on the top 1K predictions and {\sc DistMult-tanh-EV-init} tends to outperform the other methods as more predictions are generated. We think that the fact that {\sc Bilinear} starts to show more advantage over {\sc DistMult} in extracting longer rules confirm the limitation of representing relations by diagonal matrices, as longer rules requires the modeling of more complex relation semantics.

\section{Conclusion}
\label{sec:conclusion}
In this paper, we present a general framework for learning representations of entities and relations in KBs. Under the framework, we empirically evaluate different embedding models on knowledge inference tasks. We show that a simple formulation of bilinear model can outperform the state-of-the-art embedding models for link prediction on Freebase. Furthermore, we examine the learned embeddings by utilizing them to extract logical rules from KBs. We show that embeddings learned from the bilinear objective can capture compositional semantics of relations and be successfully used to extract Horn rules that involve compositional reasoning. For future work, we aim to exploit deep structure in the neural network framework. As learning representations using deep networks has shown great success in various applications \citep{Hinton2012,Vinyals12,deng2013new}, it may also help capturing hierarchical structure hidden in the multi-relational data. Further, tensor constructs have been usefully applied to some deep learning architectures \citep{Yu2013,HutchinsonPAMI}. Related constructs and architectures may help improve multi-relational learning and inference. 


\section*{Appendix}
\renewcommand{\thesubsection}{\Alph{subsection}}

\subsection{Examples of the extracted Horn rules}
Examples of length-2 rules extracted by {\sc EmbedRule} with embeddings learned from {\sc DistMult}-tanh-EV-init:
$$AwardInCeremany(a, b) \land CeremanyEventType(b, c) \implies AwardInEventType(a,c)$$
$$AtheletePlayInTeam(a, b) \land TeamPlaySport(b, c) \implies AtheletePlaySport(a, c)$$
$$TVProgramInTVNetwork(a, b) \land TVNetworkServiceLanguage(b, c) \implies TVProgramLanguage(a, c)$$
$$LocationInState(a, b) \land StateInCountry(b, c) \implies LocationInCountry(a, c)$$
$$BornInLocation(a, b) \land LocationInCountry(b, c) \implies Nationality(a, c)$$

Examples of length-3 rules extracted by {\sc EmbedRule} with embeddings learned from {\sc DistMult}-tanh-EV-init:
$$SportPlayByTeam(a, b) \land TeamInClub(b, c) \land ClubHasPlayer(c, d) \implies SportPlayByAthelete(a, d)$$
$$MusicTrackPerformer(a, b) \land PeerInfluence(b, c) \land PerformRole(c, d) \implies MusicTrackRole(a, d)$$
$$FilmHasActor(a, b) \land CelebrityFriendship(b, c) \land PersonLanguage(c, d) \implies FilmLanguage(a, d)$$

\subsection{Visualization of the relation embeddings}
\begin{figure}[ht]
\centering
\includegraphics[width=.7\linewidth]{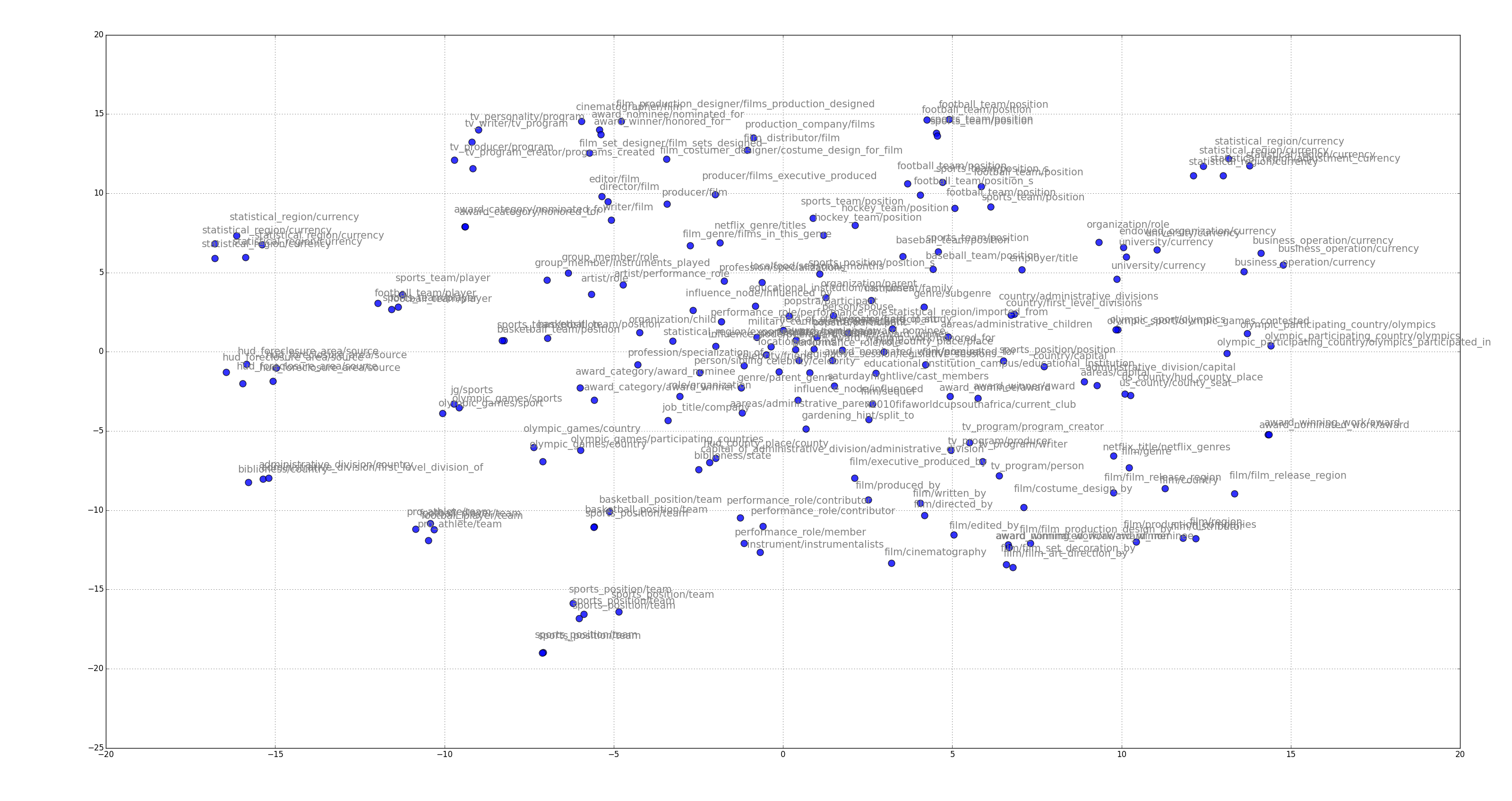}
\caption{Relation embeddings ({\sc DistAdd})}
\label{fig:distadd}
\end{figure}

\begin{figure}[ht]
\centering
\includegraphics[width=.7\linewidth]{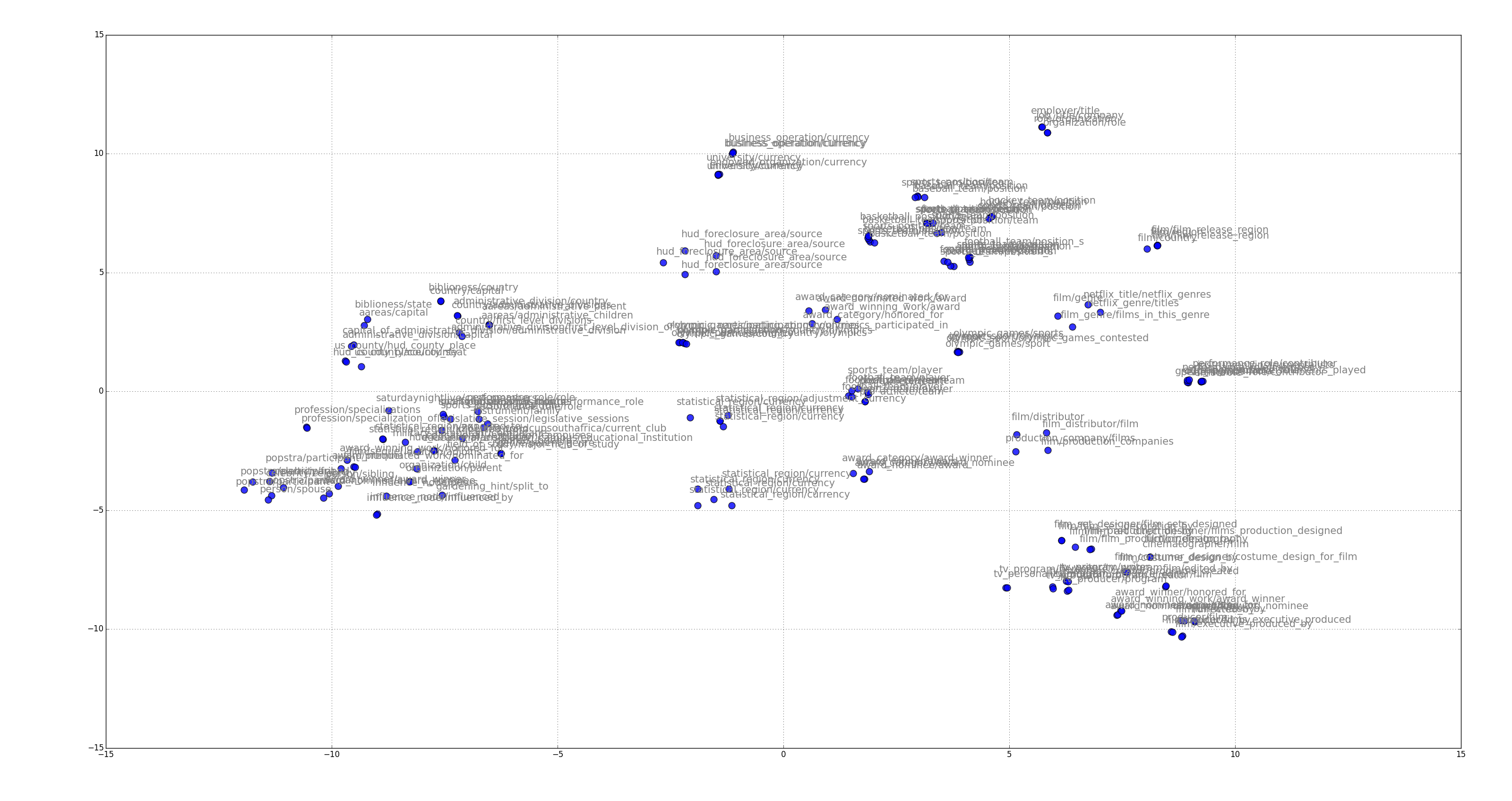}
\caption{Relation embeddings ({\sc DistMult})}
\label{fig:distmult}
\end{figure}

Visualization of the relation embeddings learned by {\sc DistMult} and {\sc DistAdd} using t-SNE (see figure~\ref{fig:distadd} and~\ref{fig:distmult}). We selected $189$ relations in the FB15k-401 dataset. The embeddings learned by {\sc DistMult} nicely reflect the clustering structures among these relations (e.g. /film/release\_region is closed to /film/country); whereas the embeddings learned by {\sc DistAdd} present structure that is harder to interpret. 

\bibliography{ref}

\begin{thebibliography}{35}
\providecommand{\natexlab}[1]{#1}
\providecommand{\url}[1]{\texttt{#1}}
\expandafter\ifx\csname urlstyle\endcsname\relax
  \providecommand{\doi}[1]{doi: #1}\else
  \providecommand{\doi}{doi: \begingroup \urlstyle{rm}\Url}\fi

\bibitem[Auer et~al.(2007)Auer, Bizer, Kobilarov, Lehmann, Cyganiak, and
  Ives]{auer2007dbpedia}
Auer, S{\"o}ren, Bizer, Christian, Kobilarov, Georgi, Lehmann, Jens, Cyganiak,
  Richard, and Ives, Zachary.
\newblock Dbpedia: A nucleus for a web of open data.
\newblock In \emph{The semantic web}, pp.\  722--735. Springer, 2007.

\bibitem[Bordes et~al.(2011)Bordes, Weston, Collobert, and
  Bengio]{bordes2011learning}
Bordes, Antoine, Weston, Jason, Collobert, Ronan, and Bengio, Yoshua.
\newblock Learning structured embeddings of knowledge bases.
\newblock In \emph{AAAI}, 2011.

\bibitem[Bordes et~al.(2013{\natexlab{a}})Bordes, Glorot, Weston, and
  Bengio]{bordes2013energy}
Bordes, Antoine, Glorot, Xavier, Weston, Jason, and Bengio, Yoshua.
\newblock A semantic matching energy function for learning with
  multi-relational data.
\newblock \emph{Machine Learning}, pp.\  1--27, 2013{\natexlab{a}}.

\bibitem[Bordes et~al.(2013{\natexlab{b}})Bordes, Usunier, Garcia-Duran,
  Weston, and Yakhnenko]{BordesUsGaWeYa2013}
Bordes, Antoine, Usunier, Nicolas, Garcia-Duran, Alberto, Weston, Jason, and
  Yakhnenko, Oksana.
\newblock Translating embeddings for modeling multi-relational data.
\newblock In \emph{NIPS}, 2013{\natexlab{b}}.

\bibitem[Bowman(2014)]{Bowman2014}
Bowman, Samuel~R.
\newblock Can recursive neural tensor networks learn logical reasoning?
\newblock In \emph{ICLR}, 2014.

\bibitem[Chang et~al.(2014)Chang, Yih, Yang, and Meek]{CYYM14}
Chang, Kai-Wei, Yih, Wen-tau, Yang, Bishan, and Meek, Chris.
\newblock Typed tensor decomposition of knowledge bases for relation
  extraction.
\newblock In \emph{EMNLP}, 2014.

\bibitem[Deng et~al.(2013)Deng, Hinton, and Kingsbury]{deng2013new}
Deng, Li, Hinton, G., and Kingsbury, B.
\newblock {New types of deep neural network learning for speech recognition and
  related applications: An overview}.
\newblock In \emph{in ICASSP}, 2013.

\bibitem[Duchi et~al.(2011)Duchi, Hazan, and Singer]{duchi2011adaptive}
Duchi, John, Hazan, Elad, and Singer, Yoram.
\newblock Adaptive subgradient methods for online learning and stochastic
  optimization.
\newblock \emph{The Journal of Machine Learning Research}, 12:\penalty0
  2121--2159, 2011.

\bibitem[Gal{\'a}rraga et~al.(2013)Gal{\'a}rraga, Teflioudi, Hose, and
  Suchanek]{galarraga2013amie}
Gal{\'a}rraga, Luis~Antonio, Teflioudi, Christina, Hose, Katja, and Suchanek,
  Fabian.
\newblock Amie: association rule mining under incomplete evidence in
  ontological knowledge bases.
\newblock In \emph{WWW}, 2013.

\bibitem[Gao et~al.(2014)Gao, Pantel, Gamon, He, Deng, and Shen]{Gao-2014}
Gao, Jianfeng, Pantel, Patrick, Gamon, Michael, He, Xiaodong, Deng, Li, and
  Shen, Yelong.
\newblock Modeling interestingness with deep neural networks.
\newblock In \emph{EMNLP}, 2014.

\bibitem[Garc{\'\i}a-Dur{\'a}n et~al.(2014)Garc{\'\i}a-Dur{\'a}n, Bordes, and
  Usunier]{garcia2014effective}
Garc{\'\i}a-Dur{\'a}n, Alberto, Bordes, Antoine, and Usunier, Nicolas.
\newblock Effective blending of two and three-way interactions for modeling
  multi-relational data.
\newblock In \emph{Machine Learning and Knowledge Discovery in Databases}, pp.\
   434--449. Springer, 2014.

\bibitem[Getoor \& Taskar(2007)Getoor and Taskar]{GetoorTa07}
Getoor, Lise and Taskar, Ben (eds.).
\newblock \emph{Introduction to Statistical Relational Learning}.
\newblock The MIT Press, 2007.

\bibitem[Grefenstette(2013)]{Grefenstette2013}
Grefenstette, Edward.
\newblock Towards a formal distributional semantics: Simulating logical calculi
  with tensors.
\newblock In \emph{*SEM}, 2013.

\bibitem[Hinton et~al.(2012)Hinton, Deng, Yu, Dahl, Mohamed, Jaitly, Senior,
  Vanhoucke, Nguyen, Sainath, and Kingsbury]{Hinton2012}
Hinton, Geoff, Deng, L., Yu, D., Dahl, G., Mohamed, A., Jaitly, N., Senior, A.,
  Vanhoucke, V., Nguyen, P., Sainath, T., and Kingsbury, B.
\newblock Deep neural networks for acoustic modeling in speech recognition.
\newblock \emph{IEEE Sig. Proc. Mag.}, 29:\penalty0 82--97, 2012.

\bibitem[Huang et~al.(2013)Huang, He, Gao, Deng, Acero, and Heck]{Huang-2013}
Huang, Po-Sen, He, Xiaodong, Gao, Jianfeng, Deng, Li, Acero, Alex, and Heck,
  Larry.
\newblock Learning deep structured semantic models for {Web} search using
  clickthrough data.
\newblock In \emph{CIKM}, 2013.

\bibitem[Hutchinson et~al.(2013)Hutchinson, Deng, and Yu]{HutchinsonPAMI}
Hutchinson, B, Deng, L., and Yu, D.
\newblock Tensor deep stacking networks.
\newblock \emph{IEEE Transactions on Pattern Analysis and Machine
  Intelligence}, 35\penalty0 (8):\penalty0 1944--1957, 2013.

\bibitem[Jenatton et~al.(2012)Jenatton, Le~Roux, Bordes, and
  Obozinski]{jenatton2012latent}
Jenatton, Rodolphe, Le~Roux, Nicolas, Bordes, Antoine, and Obozinski,
  Guillaume.
\newblock A latent factor model for highly multi-relational data.
\newblock In \emph{NIPS}, 2012.

\bibitem[Kemp et~al.(2006)Kemp, Tenenbaum, Griffiths, Yamada, and
  Ueda]{kemp2006learning}
Kemp, Charles, Tenenbaum, Joshua~B, Griffiths, Thomas~L, Yamada, Takeshi, and
  Ueda, Naonori.
\newblock Learning systems of concepts with an infinite relational model.
\newblock In \emph{AAAI}, volume~3, pp.\ ~5, 2006.

\bibitem[Mikolov et~al.(2013)Mikolov, Sutskever, Chen, Corrado, and
  Dean]{mikolov2013distributed}
Mikolov, Tomas, Sutskever, Ilya, Chen, Kai, Corrado, Greg~S, and Dean, Jeff.
\newblock Distributed representations of words and phrases and their
  compositionality.
\newblock In \emph{NIPS}, pp.\  3111--3119, 2013.

\bibitem[Nickel et~al.(2011)Nickel, Tresp, and Kriegel]{NickelTrKr11}
Nickel, Maximilian, Tresp, Volker, and Kriegel, Hans-Peter.
\newblock A three-way model for collective learning on multi-relational data.
\newblock In \emph{ICML}, pp.\  809--816, 2011.

\bibitem[Nickel et~al.(2012)Nickel, Tresp, and Kriegel]{nickel2012factorizing}
Nickel, Maximilian, Tresp, Volker, and Kriegel, Hans-Peter.
\newblock Factorizing {YAGO}: scalable machine learning for linked data.
\newblock In \emph{WWW}, pp.\  271--280, 2012.

\bibitem[Paccanaro \& Hinton(2001)Paccanaro and Hinton]{paccanaro2001learning}
Paccanaro, Alberto and Hinton, Geoffrey~E.
\newblock Learning distributed representations of concepts using linear
  relational embedding.
\newblock \emph{IEEE Transactions on Knowledge and Data Engineering},
  13\penalty0 (2):\penalty0 232--244, 2001.

\bibitem[Richardson \& Domingos(2006)Richardson and
  Domingos]{richardson2006markov}
Richardson, Matthew and Domingos, Pedro.
\newblock Markov logic networks.
\newblock \emph{Machine learning}, 62\penalty0 (1-2):\penalty0 107--136, 2006.

\bibitem[Rockt{\"a}schel et~al.(2014)Rockt{\"a}schel, Bo\v{s}njak, Singh, and
  Riedel]{Tim2014}
Rockt{\"a}schel, Tim, Bo\v{s}njak, Matko, Singh, Sameer, and Riedel, Sebastian.
\newblock Low-dimensional embeddings of logic.
\newblock In \emph{ACL Workshop on Semantic Parsing}, 2014.

\bibitem[Schoenmackers et~al.(2010)Schoenmackers, Etzioni, Weld, and
  Davis]{schoenmackers2010learning}
Schoenmackers, Stefan, Etzioni, Oren, Weld, Daniel~S, and Davis, Jesse.
\newblock Learning first-order horn clauses from web text.
\newblock In \emph{EMNLP}, 2010.

\bibitem[Shen et~al.(2014{\natexlab{a}})Shen, He, Gao, Deng, and
  Mesnil]{Shen-2014}
Shen, Yelong, He, Xiaodong, Gao, Jianfeng, Deng, Li, and Mesnil, Gregoire.
\newblock A latent semantic model with convolutional-pooling structure for
  information retrieval.
\newblock In \emph{CIKM}, 2014{\natexlab{a}}.

\bibitem[Shen et~al.(2014{\natexlab{b}})Shen, He, Gao, Deng, and
  Mesnil]{shen2014learning}
Shen, Yelong, He, Xiaodong, Gao, Jianfeng, Deng, Li, and Mesnil, Gr{\'e}goire.
\newblock Learning semantic representations using convolutional neural networks
  for {Web} search.
\newblock In \emph{WWW}, pp.\  373--374, 2014{\natexlab{b}}.

\bibitem[Singh \& Gordon(2008)Singh and Gordon]{singh2008relational}
Singh, Ajit~P and Gordon, Geoffrey~J.
\newblock Relational learning via collective matrix factorization.
\newblock In \emph{KDD}, pp.\  650--658. ACM, 2008.

\bibitem[Socher et~al.(2012)Socher, Huval, Manning, and Ng]{socher:semantic}
Socher, Richard, Huval, Brody, Manning, Christopher~D., and Ng, Andrew~Y.
\newblock Semantic compositionality through recursive matrix-vector spaces.
\newblock In \emph{EMNLP-CoNLL}, 2012.

\bibitem[Socher et~al.(2013)Socher, Chen, Manning, and
  Ng]{SocherChenManningNg2013}
Socher, Richard, Chen, Danqi, Manning, Christopher~D., and Ng, Andrew~Y.
\newblock Reasoning with neural tensor networks for knowledge base completion.
\newblock In \emph{NIPS}, 2013.

\bibitem[Suchanek et~al.(2007)Suchanek, Kasneci, and Weikum]{suchanek2007yago}
Suchanek, Fabian~M, Kasneci, Gjergji, and Weikum, Gerhard.
\newblock Yago: a core of semantic knowledge.
\newblock In \emph{WWW}, 2007.

\bibitem[Sutskever et~al.(2009)Sutskever, Tenenbaum, and
  Salakhutdinov]{sutskever2009modelling}
Sutskever, Ilya, Tenenbaum, Joshua~B, and Salakhutdinov, Ruslan.
\newblock Modelling relational data using {Bayesian} clustered tensor
  factorization.
\newblock In \emph{NIPS}, pp.\  1821--1828, 2009.

\bibitem[Vinyals et~al.(2012)Vinyals, Jia, Deng, and Darrell]{Vinyals12}
Vinyals, O., Jia, Y., Deng, L., and Darrell, T.
\newblock Learning with recursive perceptual representations.
\newblock In \emph{NIPS}, 2012.

\bibitem[Yih et~al.(2014)Yih, He, and Meek]{yihsemantic}
Yih, Wen-tau, He, Xiaodong, and Meek, Christopher.
\newblock Semantic parsing for single-relation question answering.
\newblock In \emph{ACL}, 2014.

\bibitem[Yu et~al.(2013)Yu, Deng, and Seide]{Yu2013}
Yu, D., Deng, L., and Seide, F.
\newblock The deep tensor neural network with applications to large vocabulary
  speech recognition.
\newblock \emph{IEEE Trans. Audio, Speech and Language Proc.}, 21\penalty0
  (2):\penalty0 388 --396, 2013.

\end{thebibliography}
\bibliographystyle{iclr2015}

\end{document}